# Developing FB Chatbot Based on Deep Learning Using RASA Framework for University Enquiries

**Yurio Windiatmoko, Ahmad Fathan Hidayatullah, Ridho Rahmadi,**

yurio.windiatmoko@students.uii.ac.id , fathan@uii.ac.id , ridho.rahmadi@uii.ac.id,

Department of Informatics, Universitas Islam Indonesia, Yogyakarta, Indonesia

**Abstract**. Smart systems for Universities powered by Artificial Intelligence have been massively developed to help humans in various tasks. The chatbot concept is not something new in today's society which is developing with recent technology. College students or candidates of college students often need actual information like asking for something to customer service, especially during this pandemic, when it is difficult to have an immediate face-to-face meeting. Chatbots are functionally helping in several things such as curriculum information, admission for new students, schedule info for any lecture courses, students grade information, and some adding features for Muslim worships schedule, also weather forecast information. This Chatbot is developed by Deep Learning models, which was adopted by an artificial intelligence model that replicates human intelligence with some specific training schemes. This kind of Deep Learning is based on RNN which has some specific memory savings scheme for the Deep Learning Model, specifically this chatbot using LSTM which already integrates by RASA framework. LSTM is also known as Long Short Term Memory which efficiently saves some required memory but will remove some memory that is not needed. This Chatbot uses the FB platform because of the FB users have already reached up to 60.8% of its entire population in Indonesia. Here's the chatbot only focuses on case studies at campus of the Magister Informatics FTI University of Islamic Indonesia. This research is a first stage development within fairly sufficient simulate data.

## 1. Introduction

Chatbot is the easiest system to access by any users, chatbot is an artificial intelligence technology for communicating through natural language with humans. Chatbot makes it possible to understand human language through Natural Language Processing. Language is a social and cultural reciprocal for a natural protocol. It is time for computer machines to add features for understanding humans along with the increased performance of the recent computer machine hardware [1].

College students or candidates of college students often need actual information like asking for something to customer service, especially during this pandemic, when it is difficult to have an immediate face-to-face meeting. Therefore, a chatbot is needed to answer every question quickly, correctly, and is available anytime which is expected to work for 24 hours [2]. Functionally this Chatbot is helping in several things such as curriculum information, admission for new students, schedule info for any lecture courses, students grade information, and some adding features for Muslim worships schedule, also weather forecast information.

This Chatbot is developed by Deep Learning models, which was adopted by an artificial intelligence model that replicates human intelligence with some specific training schemes. Deep Learning consists of high-level abstract modeling algorithms on data using transformation of nonlinear functions arranged by many layers and deeply [3]. Deep learning is also a branch of Machine Learning, inspired by the neural network of the human brain which is represented by layers on the neural network [4].

This kind of Deep Learning is based on RNN which has some specific memory savings scheme for the Deep Learning Model, specifically this chatbot using LSTM which already integrates by RASA framework. LSTM is also known as Long Short Term Memory which efficiently saves some required memory but will remove some memory that is not needed. The implementation of Deep Learning on chatbots requires the role of the chatbot framework which is for unifying and pipelining the Deep Learning model. The sense of a chatbot framework is to implement its conversation flow. RASA works on two main procedures namely RASA NLU and RASA Core. Rasa is a library of tools from the Python programming language that is open source for building conversational software. The aim and philosophy of the initiators of the Rasa chatbot Framework are to create Machine learning-based dialogue management that provides ease of use in terms of implementation, as well as bootstrapping, by starting the stages from existing sources to create something more complex and in a more efficient way, even with minimal initial training data [5].

## 2.     Literature Review

There are two main approaches to generating a chatbot response. If the chatbot uses a traditional approach then it is hard-coded rule-based templates and rules that process for responses. However, nowadays many new and interesting approaches allow deep learning approaches to emerging. Neural network models are trained with several data to study the process of generating responses that are grammatically relevant and in terms of user speech intent [6].

These are some previous works relating to Chatting System or chatbot for universities. Online Chatting System for college inquiries using a knowledgeable database by Bathe, Malusare, and Kolpe is doing pattern matching to perform information retrieval on chatbots. It hasn't already used the Deep Learning approach, so that is still too rigid by rules for users to make inquiries. But, detailed working steps in this research are very good, there such a UML, and various kinds of process diagrams [7]. Erasmus the AI chatbot, Erasmus is a chatbot on Facebook which is used to answer the queries related to the college information. Designing end-to-end systems using cloud services, starting from api.ai (Dialogflow), Mlab (MongoDB cloud), IBM Bluemix (webhook API), scrapper import.io to minimize the scripting code process, but what's happening here, it's all because there are too many diverse cloud services, there is quite a long latency between the services [8]. Then, Eaglebot a Chatbot Based Multi-Tier Question Answering System for Retrieving Answers from Heterogeneous Sources Using BERT. It is a scalable chatbot system using 3 route selection methods with the main framework using Dialogflow, then several completion methods from document retrieval and document reader [9]. Eaglebot which serves to answer various questions commonly asked by students in the university domain. But Eaglebot still has some limitations by Dialogflow chatbot framework which has some request limits, if it wants to unlimit requests you need to subscribe for the member privileged.

Intelligent Chatbot System Based on Entity Extraction Using RASA NLU and Neural Network [10] by Anran Jiao was comparing performance between RASA NLU stack with Neural Network Classifier and Entity Extractor from scratch. This research was told that the RASA NLU method still superior to extract all the entities and classify user speech intent. It's also quite comprehensive research but this research only depends on a free API for response completion and not using its own database. And now comparing from all those previous work, this research will be more reliable by using Deep learning for chatbot natural language understanding and combining free API with it's own database for response completion. So it won't need for pattern matching, not using cloud provider service-based, and not only depends on a free API service so that customization will be easier, scalable and no limit request for user interaction.

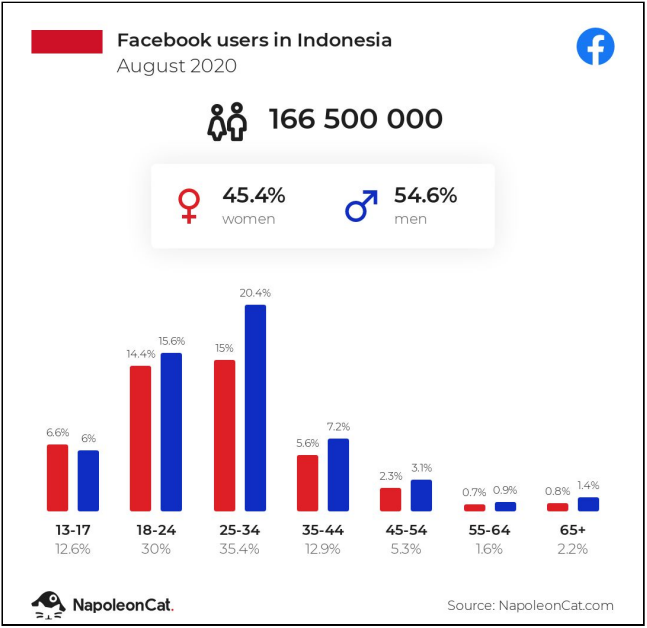

**Figure 1.** FB Users In Indonesia[1]

Facebook users in Indonesia August 2020. There were 166.500.000 Facebook users in Indonesia in August 2020, which accounted for 60.8% of its entire population. The majority of them were men - 54.6%. People aged 25 to 34 were the largest user group 59.000.000. The highest difference between men and women occurs within people aged 25 to 34, where men lead by 9.000.000. So that this Chatbot uses the FB platform as the user interaction medium.

---

[1] "Facebook users in Indonesia - August 2020 | NapoleonCat."
https://napoleoncat.com/stats/facebook-users-in-indonesia/2020/08. Accessed 16 Sep. 2020.

## 3. Data and Methods

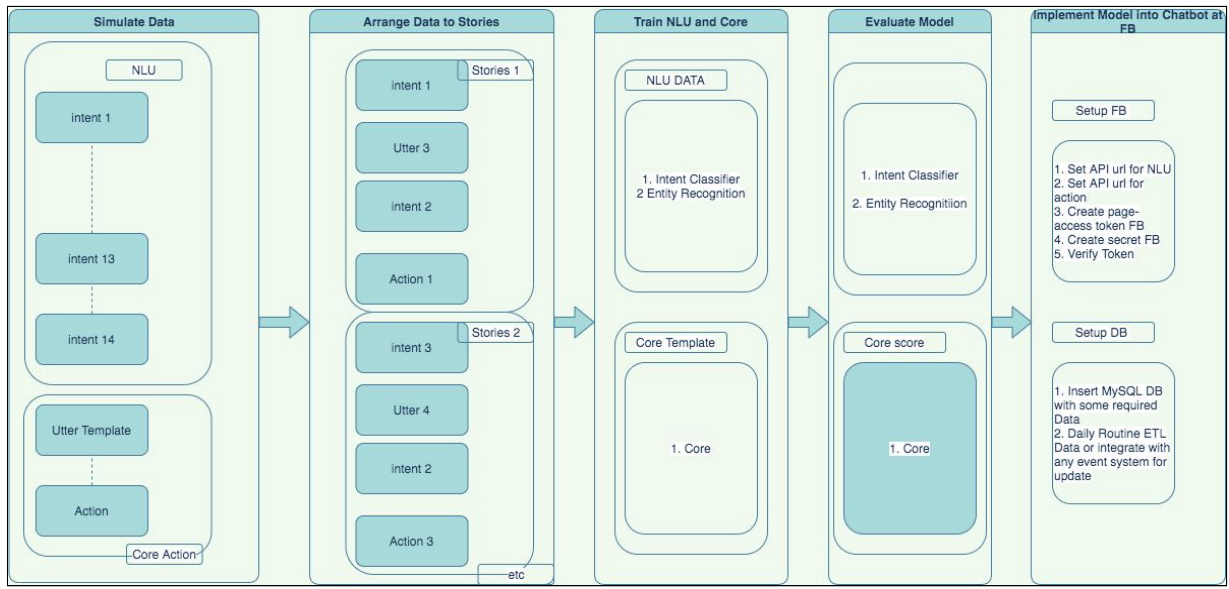

**Figure 2.** Workflow Chart Chatbot

The picture above describes flow and methods of the research process carried out. Beginning with simulating the data based on common dialogue interaction along with grouping for each sentence into the intent topic, utter chatbot response template, and action. Especially in the intent topic which has been proposed as the featured chatbot also needs defining Entities there. The difference between utter and action here is basically that utter is just a sentence with an expression of the general response, whereas action is utilized when it's response needs some completion based on a database or API, also it needs wrapping-up as the class actions on RASA. At this stage, various data previously seen and considered from general reference schemes or conversation scenarios for each chatbot feature such as communication flow on student questions about the curriculum, admission for new students, schedule for lectures, student grade, Moslem Worships schedule, and weather forecast also adding up with some common dialogue intent like greetings and so on. Then, arrange those intent, utter and action in such a way that it corresponds to the flow of stories and common entities in each conversation sentence. After all the data is already set up, then train those data using RASA Trainer. Check and recheck after training was finished by evaluating the trained model. If the evaluation score obtained has reached the required expectations then just integrate that model with credential setup configuration after the fb developer account has been created.

*3.1.    Framework*

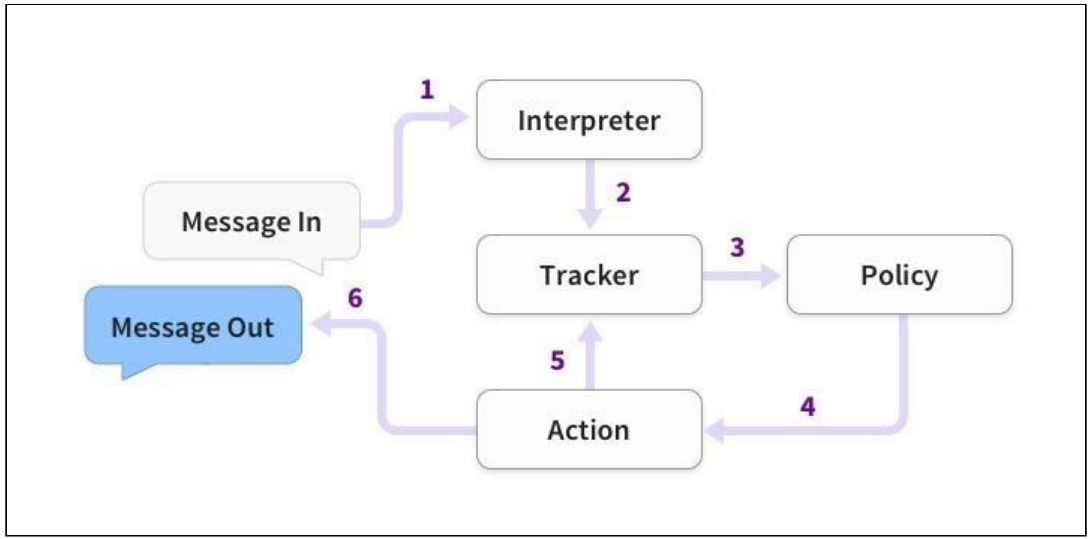

**Figure 3.** RASA Stack Framework

The architecture or dialogue management flow used in this research approach is as shown in Figure 3 which is the architecture of the chatbot stack framework itself. The first stage of the message is received and forwarded to the interpreter, namely Rasa NLU to extract intents, entities, and other structured information. Both interpreter or tracker is tracking, detecting, and maintain the status of the conversation context through the message notifications it has received. The third policy or policy manager receives context status from the tracker. The fourth policy or policymakersMuslim choose which action will be taken next. The five actions or actions are recorded by the tracker. These six actions are executed by sending a message to the user. Seventh, if the action that has been executed is ignored or ignored by the user at a certain time, the process returns to the third step.

*3.2.    Methods For Training*

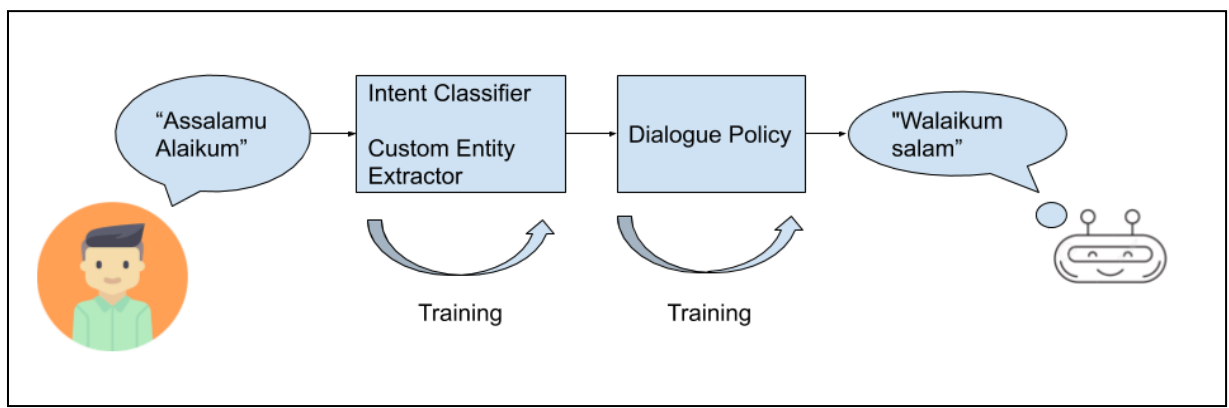

**Figure 4.** High Overview Chat System and Cycle

The above picture describes how each module is trained separately, those modules are NLU and Dialogue Policy or Dialogue Management. NLU or the interpreter comprises two component models inside. Those are Intent Classifier and Entity Extractor.

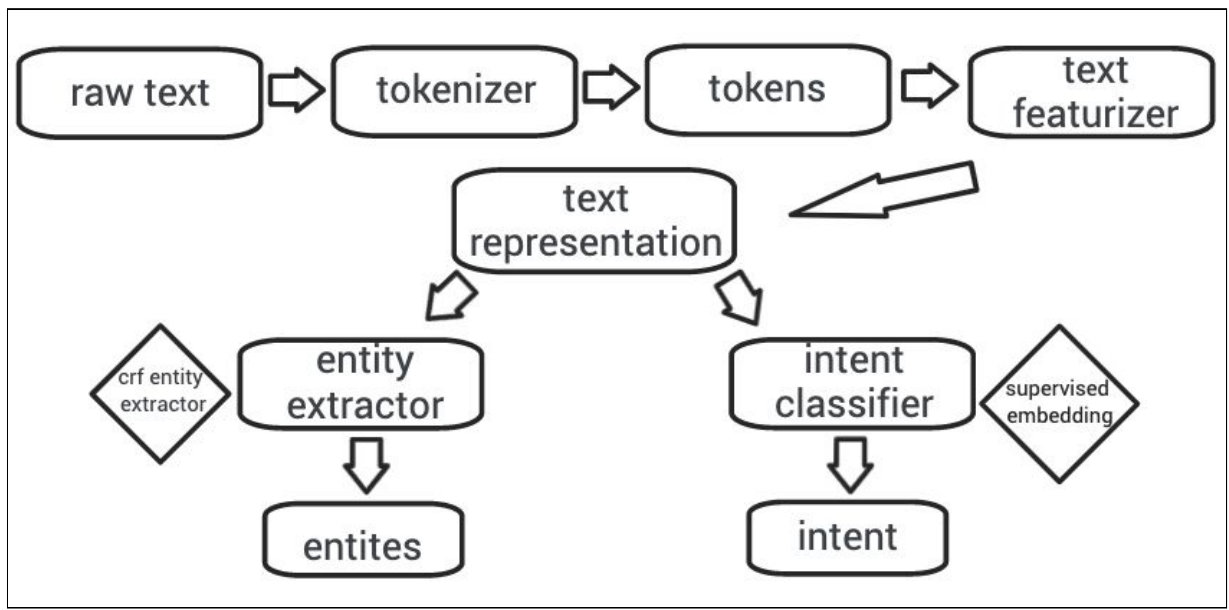

**Figure 5.** Intent Entity Training Flow

Specifically for this chatbot used Supervised Embeddings as Intent Classifier, it is also known as Embedding Intent Classifier. The Embedding Intent Classifier embeds user inputs and intent labels into the same space. Supervised embeddings are trained by maximizing similarity between them. This algorithm is based on StarSpace[2]. However, in this implementation the loss function is slightly different and additional hidden layers are added together with dropout. This algorithm also provides similarity rankings of the labels.

The training configuration to build this intent classifier model is 300 for epochs, then for the hidden layers sizes at feed forward neural networks are 256 at h1, 128 at h2 , and 14 at output neural network label intent class. Embedding dimension used for training is 20, this is multiple embedding layers inside the model architecture. For example, vector of '__CLS__' token and it's intent is passed on to an embedding layer before they are compared and the loss is calculated. config for batch size is (64, 256) which the initial and final value for batch sizes. Batch size will be linearly increased for each epoch. Learning rate is set as 0.001 for this training config.

Entity Extractor used for this chatbot is crf entity extraction[3] which method is suitable for custom entities extractor. It's component implements a conditional random field (CRF) to do named entity recognition. CRFs can be thought of as an undirected Markov chain where the time steps are words and the states are entity classes. It's Features are from the words like capitalization, POS tagging, etc. Those Features are give probabilities to certain entity classes, as are transitions between neighbouring entity tags, the most likely set of tags is then calculated and returned. Training configuration for CRFEntityExtractor has some list of features to use. However, it can be set and chosen by any requirements. Specifically this chatbot uses a list of features in such 'lowercase' to check whether the token is lower case, 'uppercase' to check whether the token is upper case, 'title' to check whether the token starts with an uppercase character and all remaining characters are lowercase, 'digit' to checks whether the token contains just digits, 'prefix5' to take the first five characters of the token, 'prefix2' to take the first two characters of the token, 'suffix5' to take the last five characters of the token, 'suffix3' to take the last three characters of the token, 'suffix2' to take the last two characters of the token, 'suffix1' to take the last character of the token, 'bias' to add an additional "bias" feature to the list of features.

---

[2] "StarSpace: Embed All The Things!." 12 Sep. 2017, https://arxiv.org/abs/1709.03856. Accessed 19 Sep. 2020.
[3] "Conditional Random Fields: Probabilistic Models for ...." 28 Jun. 2001, http://cs.nju.edu.cn/daixinyu/CRF.pdf. Accessed 21 Sep. 2020.

As the featurizer is moving over the tokens in a user message with a sliding window, it defines features for previous tokens, the current token, and the next tokens in the sliding window. so the features are as [before, token, after] the array.

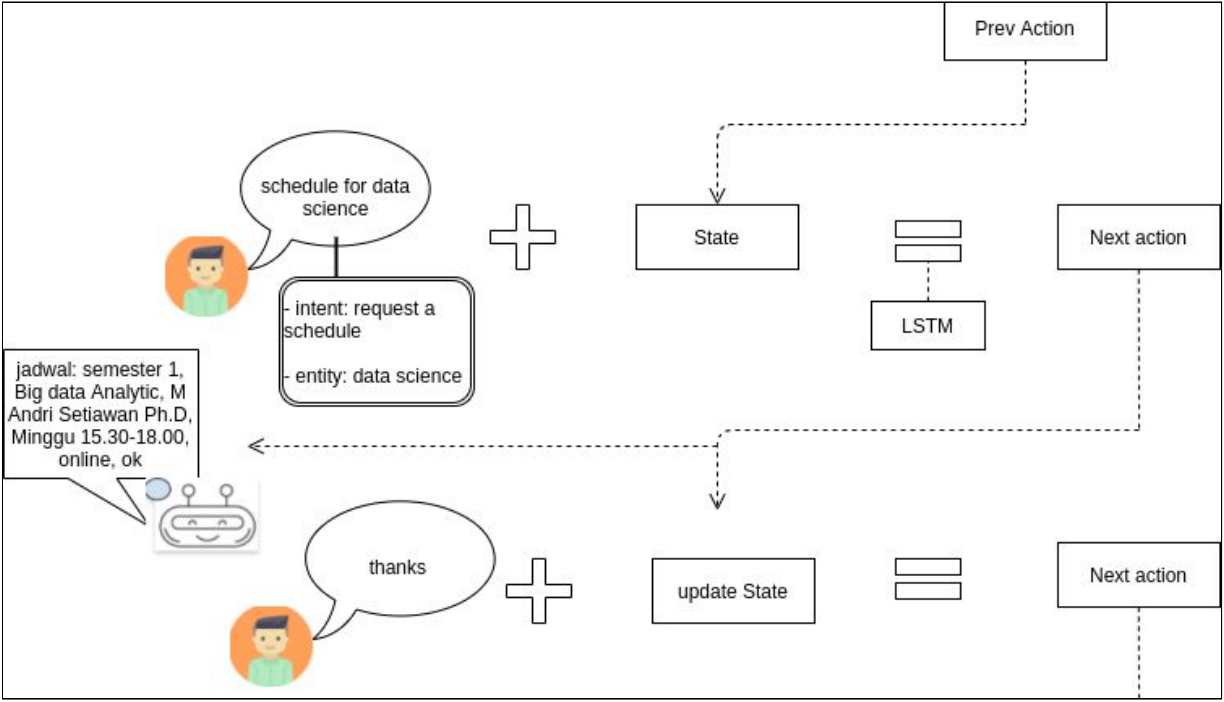

**Figure 6.** Dialogue Management Component Flow

Then next for the Dialogue Policy or Dialogue Management is works by creating training data from the stories and training a model on that data. Stories are like some scenario or scheme from the common topic in any chat flow that we design or we expected. Chat Response works by creating training data from the stories and training a model on that data like at the Figure 6 already described. Dialogue Policie decides which action to take at every step in the conversation. There are different policies to choose from, and it includes multiple policies in a single Agent or chatbot. At every turn, the policy which predicts the next action with the highest confidence will be used.

This chatbot already configured the policies by memoization Policy and neural network Policy. The execution of the policy will be based on the priority; highest priority policy will be executed first and so on. Priority is calculated on the basis of confidence score of each policy, which has a higher score given higher priority.

The MemoizationPolicy just memorizes the conversations in training data. It predicts the next action with confidence 1.0 if this exact conversation exists in the training data, otherwise it predicts None with confidence 0.0.

A Neural Network Policy implemented in Keras to select the next action. The default architecture is based on an LSTM[4]. It is configured with masking layer (input, 5, 32), lstm units (32) dropout=0.2, dense layer (19) then 100 training epochs.

*3.3.    Connector Module Platform*

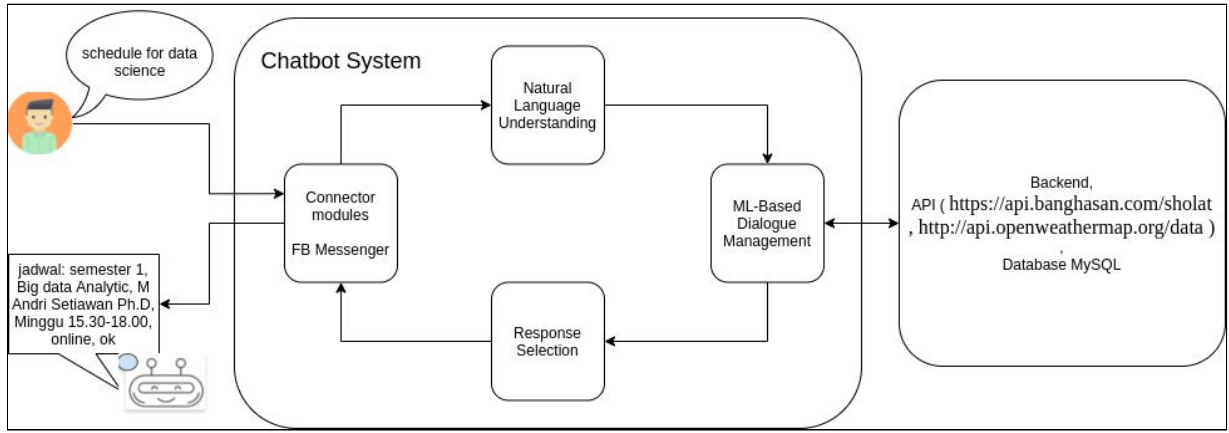

**Figure 7.** Flow of Chat System

Facebook Messenger as a connector module platform, first it needs to set up a facebook page and app to get credentials to connect to Facebook Messenger. Once already have all of it, add these to the credentials.yml file. Then set up a Webhook and select at least the messaging and messaging_postback subscriptions. Insert the callback URL which will look like https://<HOST>/webhooks/facebook/webhook. Insert and Verify Token which has to match the verify entry in the credentials.yml file. So then, webhooks allows the chatbot to receive real-time HTTP notifications of changes to specific objects and postback to messaging.

## 4.    Result and Discussion

This chatbot development research that uses deep learning models, it is expected to provide the best performance for chatbots systems through this deep learning model. Furthermore, the representation results of the performance evaluation of the deep learning model on the chatbot, namely intent classification and entity extraction at NLU and Dialogue Policy on the dialogue management chatbot system are measured and evaluated by precision score, recall score and F1-score.

Precision may be defined as the probability that an object is relevant given that it is returned by the system or here as the model predicted, while the recall is the probability that a relevant object is returned based on ground truth of each label data. Then, F1 Score is needed for seeking a balance between Precision and Recall.

---

[4] "LONG SHORT-TERM MEMORY 1 INTRODUCTION."
https://www.bioinf.jku.at/publications/older/2604.pdf. Accessed 21 Sep. 2020.

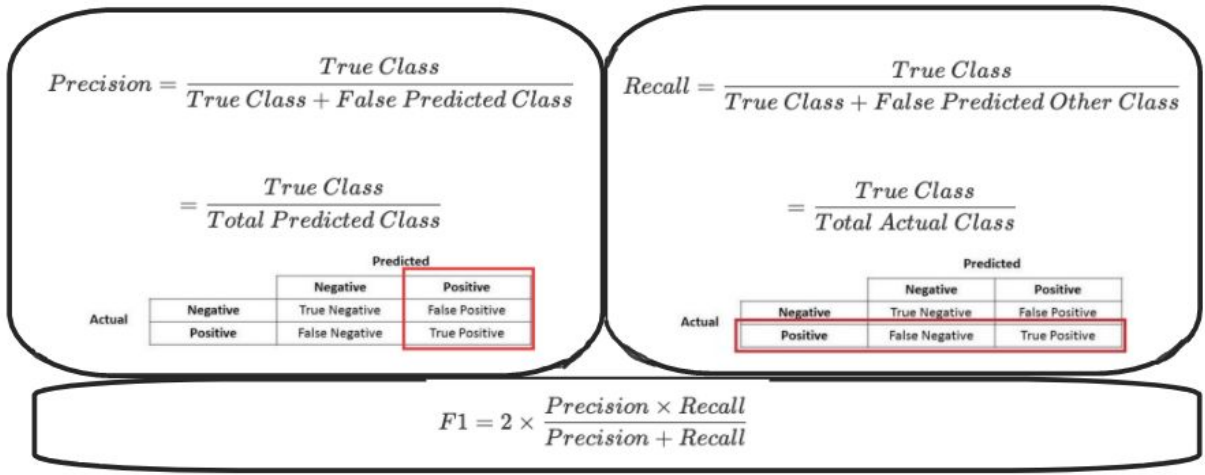

**Figure 8.** Precision Recall and F1 score Formula

Immediately, that Precision talks about how precise or accurate the model is out of those predicted Class, how many of them are actual Class. Then, Recall actually calculates how many of the Actual Class our model captures through labeling it as those Class. Thus F1 Score might be a better measure to use if it needs to seek a balance between Precision and Recall and there is an uneven class distribution.

**Table 1** Metrics Evaluation Intent Classifier 1

|  | intent agreed | intent requests a schedule | intent goodbye | intent Muslim greeting | mood unhappy | intent only request a schedule | intent weather forecast | intent request a worship schedule |
|---|---|---|---|---|---|---|---|---|
| precision | 1 | 1 | 1 | 0.75 | 1 | 1 | 1 | 1 |
| recall | 1 | 1 | 1 | 1 | 1 | 1 | 1 | 1 |
| f1-score | 1 | 1 | 1 | 0.85 | 1 | 1 | 1 | 1 |
| data count | 6 | 100 | 3 | 3 | 3 | 4 | 5 | 11 |
| confused with | {} | {} | {} | {} | {} | {} | {} | {} |

Here what needs to be underlined is the Intent Muslim greeting class where the model can predict accurately to the amount of truth data for that class which the recall score is 100%, but the model still predicts the intent Muslim greeting in other intent classes actual which the precision score is 75%, namely the intent greeting class.

**Table 2** Metrics Evaluation Intent Classifier 2

| | mood happy | intent greeting | intent thanks | intent request grade score | intent refuse | challenge bot | avg | total |
|---|---|---|---|---|---|---|---|---|
| precision | 1 | 1 | 1 | 1 | 1 | 1 | 0.982 | 13.75 |
| recall | 1 | 0.85 | 1 | 1 | 1 | 1 | 0.98 | 13.85 |
| f1-score | 1 | 0.92 | 1 | 1 | 1 | 1 | 0.98 | 13.77 |
| data count | 4 | 7 | 2 | 10 | 3 | 4 | 11.78 | 165 |
| confused with | {} | {'intent Muslim greeting: 1} | {} | {} | {} | {} | - | - |

While for the intent greeting itself, here is get a 100% precision score where the Intent greeting is sometimes not detected by the model on the contrary intent Muslim greeting is even detected because it uses the same word phrase, which is 'assalamu alaikum' but this is not a problem because the chatbot template utter will answer the same thing as greeting utter.

**Table 3** Confusion Matrix Between Intent Muslim Greeting and Intent Greeting

| | | predicted | | | |
|---|---|---|---|---|---|
| | | intent Muslim greeting | intent greeting | data count | recall |
| actual | intent Muslim greeting | 3 | 0 | 3 | 1 |
| | intent greeting | 1 | 6 | 7 | 0.85 |
| | total predicted | 4 | 6 | f1-score | |
| | precision | 0.75 | 1 | 0.85 | 0.92 |

The table above is a detailed explanation of the precision and recall calculations of the two classes that are often misclassified

**Table 4** Metrics Evaluation Entity Recognition

| | name | concentration study program | city | NIM | avg | total |
|---|---|---|---|---|---|---|
| precision | 1 | 1 | 1 | 1 | 1 | 4 |
| recall | 1 | 1 | 1 | 1 | 1 | 4 |
| f1-score | 1 | 1 | 1 | 1 | 1 | 4 |

| | data count | 10 | 170 | 12 | 3 | 48.75 | 195 |
|---|---|---|---|---|---|---|---|

The entity class metrics evaluation table has shown the best performance from the model to be able for extracting the entity labels.

**Table 5** Confusion Matrix Dialogue Policy Decision

| | action schedule list | action listen | utter did it help | utter good bye | utter salam replies | utter say hello | utter cheers | utter im bot | utter happy | utter asked concentration study program | utter thank you |
|---|---|---|---|---|---|---|---|---|---|---|---|
| action schedule list | 4 | 0 | 0 | 0 | 0 | 0 | 0 | 0 | 0 | 0 | 0 |
| action listen | 0 | 27 | 0 | 0 | 0 | 0 | 0 | 0 | 0 | 0 | 0 |
| utter did it help | 0 | 0 | 2 | 0 | 0 | 0 | 0 | 0 | 0 | 0 | 0 |
| utter goodbye | 0 | 0 | 0 | 2 | 0 | 0 | 0 | 0 | 0 | 0 | 0 |
| utter salam replies | 0 | 0 | 0 | 0 | 3 | 0 | 0 | 0 | 0 | 0 | 0 |
| utter say hello | 0 | 0 | 0 | 0 | 0 | 6 | 0 | 0 | 0 | 0 | 0 |
| utter cheers | 0 | 0 | 0 | 0 | 0 | 0 | 2 | 0 | 0 | 0 | 0 |
| utter im bot | 0 | 0 | 0 | 0 | 0 | 0 | 0 | 1 | 0 | 0 | 0 |
| utter happy | 0 | 0 | 0 | 0 | 0 | 0 | 0 | 0 | 2 | 0 | 0 |
| utter asked concentration study program | 0 | 0 | 0 | 0 | 0 | 0 | 0 | 0 | 0 | 2 | 0 |
| utter thank you | 0 | 0 | 0 | 0 | 0 | 0 | 0 | 0 | 0 | 0 | 5 |

Likewise, the dialogue policy performance has shown the best performance of the model in making decisions for the chatbot response.

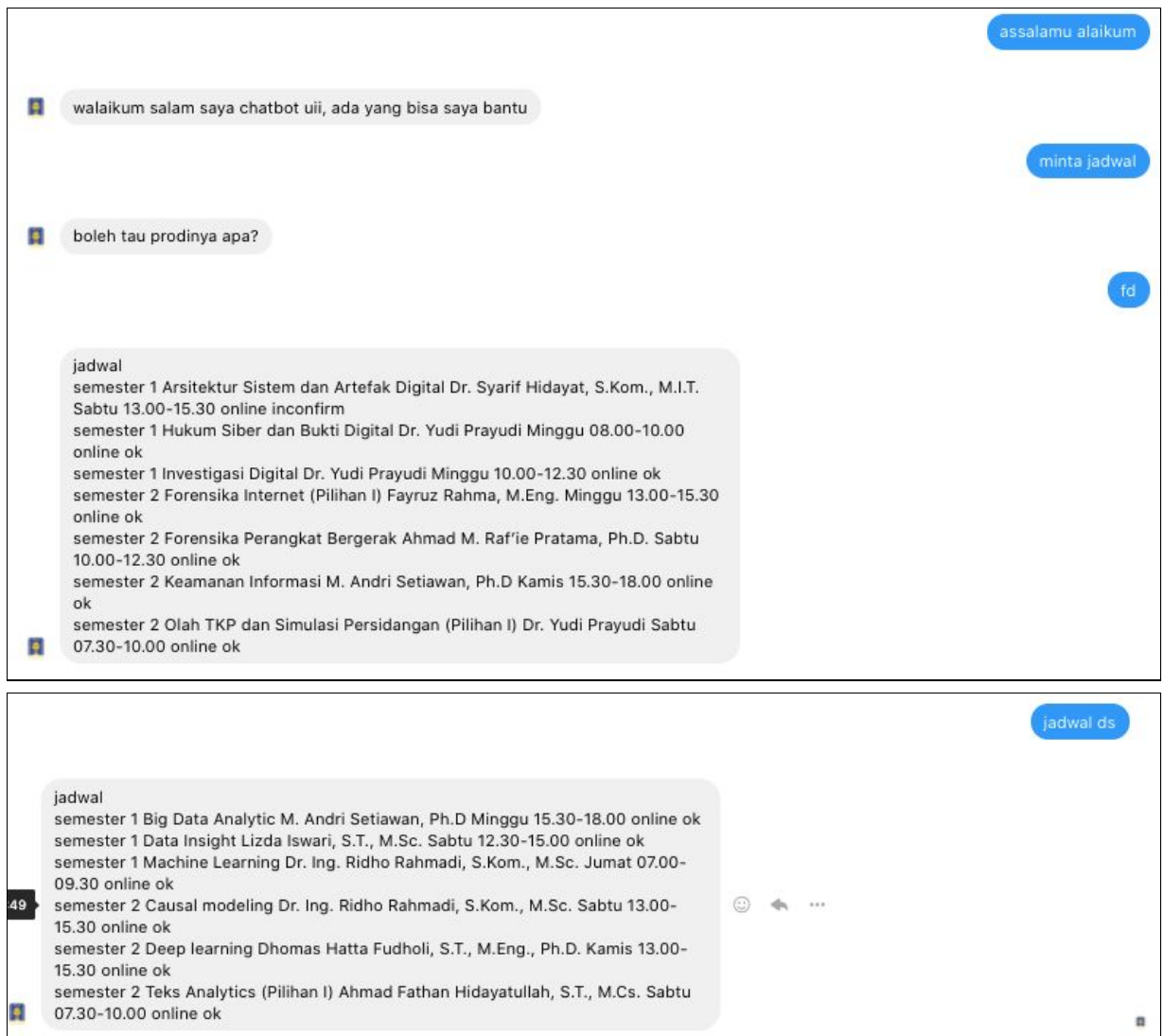

**Figure 8** Chat of students requesting a schedule

Figure 8 illustrates the student conversation flow requesting a lecture schedule from the chatbot. The chat is started with the greetings made by students to the chatbot, and the chatbot detects the greeting intent from users and replies with utter replies to the greeting. Furthermore, students go to the point to ask about their lecture schedule, and Chatbot detects it as an intent to request a schedule from users, then replies with an utter asking about the concentration of the student's study program. Then the student tells the chatbot the concentration of his study program, namely 'fd' or digital forensics, and the chatbot considers this to be the concentration entity of the study program and continues the query to MySQL DB as a completion to get lecture schedule data.

As for the alternative, the process of requesting schedule also may ask directly for a specific study program concentration schedule, for example, 'schedule ds' and the chatbot will detect the intent to request the schedule as well as to detect the concentration entity of the study program 'data science'. And so chatbot continues the query to MySQL DB as a completion to get lecture schedule data.

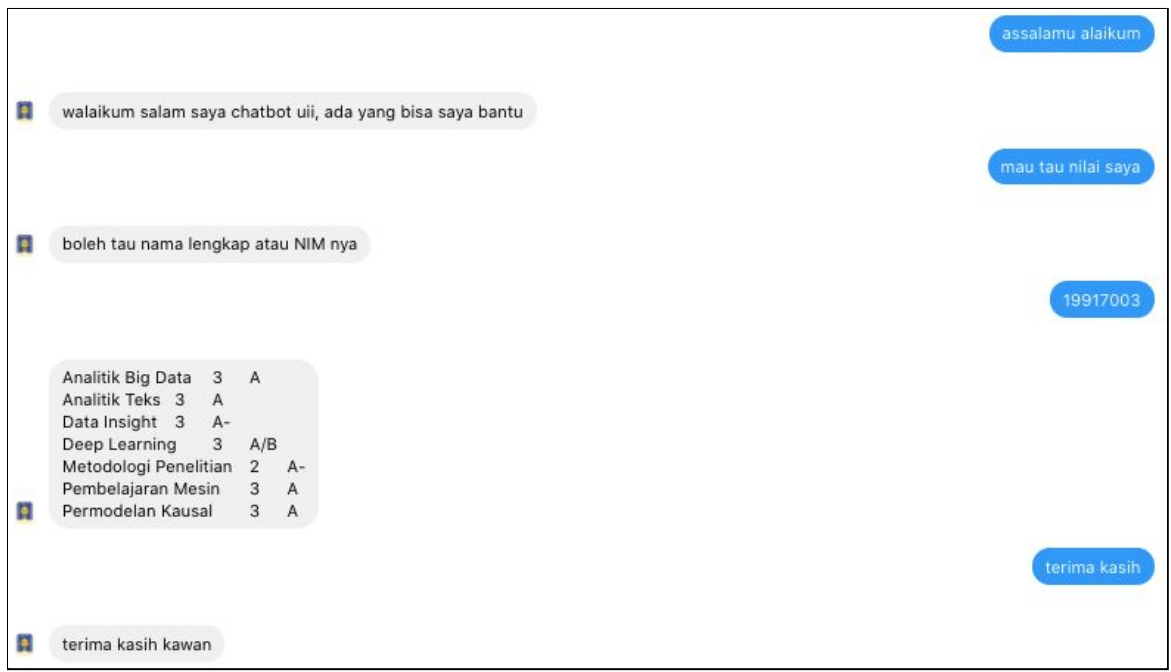

**Figure 9** Chat of students requesting grades score

Then this flow about requesting a grade score is almost the same as requesting lecture schedule flow, but only with different intents and entities, also different actions to show a list of grade scores by continuing the completion query to MySQL DB as a completion to get lecture schedule data.

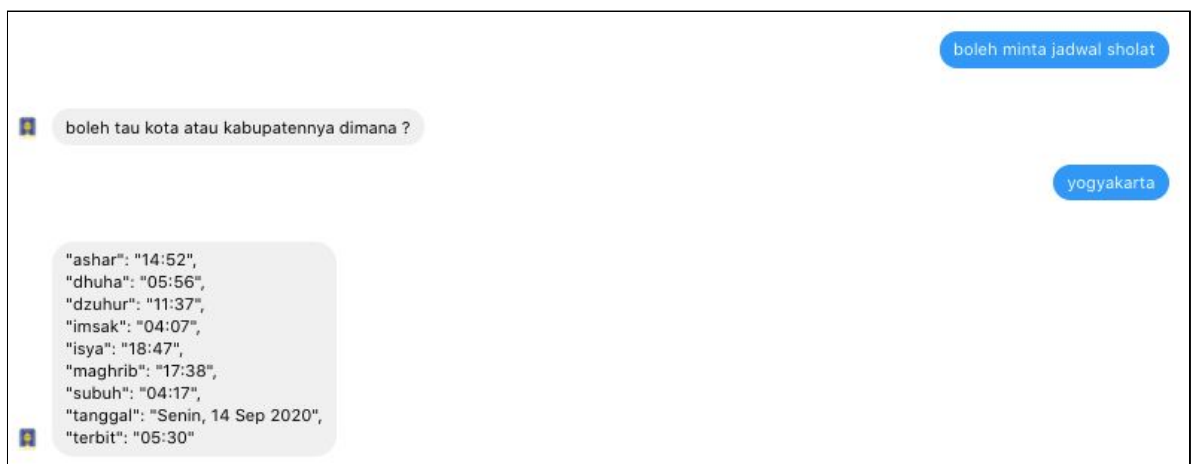

**Figure 10** Chat of students requesting Moslem worship schedule

This picture above is a conversation flow of requesting the Moslem worship schedule. Beginning by asking directly for a specific intent about asking the Moslem worship schedule. Then the chatbot asking about a specific location as an entity, after that continuing the completion query directly to API

https://api.banghasan.com/sholat/format/json/jadwal/kota/{city}/tanggal/{date.now}[5].

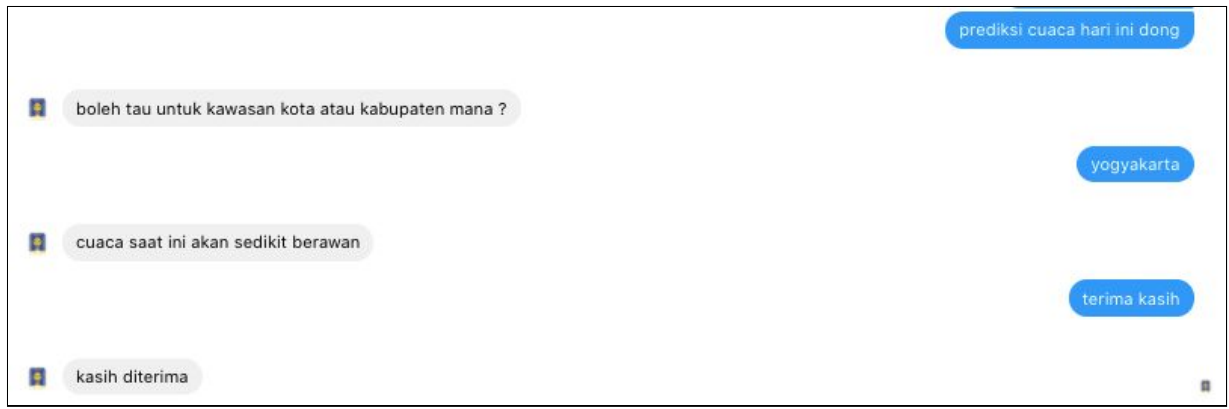

**Figure 11** Chat of students requesting weather forecast of today

Then a conversation flow of requesting a weather forecast for today. Beginning by asking directly for a specific intent about the weather forecast for today. Then the chatbot asking about a specific location as an entity, after that continuing the completion query directly to API.

http://api.openweathermap.org/data/2.5/weather?q={Yogyakarta,id}&APPID={your_app_id}&units=metric[6],

Then translate the weather forecast description into Indonesian Language.

## 5. Conclusion and Future Work

So far the chatbot has been able to provide the best performance in terms of the intent class, entity class and also the appropriate reply from the dialogue policy in the chatbot model based on the great result from each evaluation metrics.

This research is still in a first stage development within fairly sufficient simulate data, but we also have implemented deep learning models for chatbots, especially by implementing the intent classifier and entity recognition on rasa nlu also dialogue policy on rasa core, then integrating it with Facebook Messenger.

Furthermore, this chatbot needs to be developed by extending the dialogue data and more features for helping the automating task in the universities, then by directing towards the open domain chatbot, for more extensive implementation in Deep Learning Chatbot, to be the 'chatbot I know everything'. Also the needs for implementing chatbot within its own platform, no limitations then and more customizable.

---

[5] "Fathimah API." https://api.banghasan.com/. Accessed 16 Sep. 2020.
[6] "Weather API - OpenWeatherMap." https://openweathermap.org/api. Accessed 16 Sep. 2020.